# Finding Optimal Bayesian Networks

**David Maxwell Chickering**
Microsoft Research
Redmond, WA 98052
dmax@microsoft.com

**Christopher Meek**
Microsoft Research
Redmond, WA 98052
meek@microsoft.com

## Abstract

In this paper, we derive optimality results for greedy Bayesian-network search algorithms that perform single-edge modifications at each step and use asymptotically consistent scoring criteria. Our results extend those of Meek (1997) and Chickering (2002), who demonstrate that in the limit of large datasets, if the generative distribution is perfect with respect to a DAG defined over the observable variables, such search algorithms will identify this optimal (i.e. generative) DAG model. We relax their assumption about the generative distribution, and assume only that this distribution satisfies the *composition property* over the observable variables, which is a more realistic assumption for real domains. Under this assumption, we guarantee that the search algorithms identify an *inclusion-optimal* model; that is, a model that (1) contains the generative distribution and (2) has no sub-model that contains this distribution. In addition, we show that the composition property is guaranteed to hold whenever the dependence relationships in the generative distribution can be characterized by paths between singleton elements in some generative graphical model (e.g. a DAG, a chain graph, or a Markov network) even when the generative model includes unobserved variables, and even when the observed data is subject to selection bias.

## 1  Introduction

The problem of learning Bayesian networks (a.k.a directed graphical models) from data has received much attention in the UAI community. A simple approach taken by many researchers, particularly those con-tributing experimental papers, is to apply—in conjunction with a scoring criterion—a greedy single-edge search algorithm to the space of Bayesian-network structures or to the space of equivalence classes of those structures. There are a number of important reasons for the popularity of this approach. First, a variety of hardness results have shown that learning various classes of Bayesian networks is NP-hard (e.g. Chickering, 1996; Dasgupta, 1999; Meek, 2001), and it is widely accepted that heuristic search algorithms are appropriate in general. Second, greedy search is simple to implement and the evaluation of single-edge modifications is computationally efficient. Third, and perhaps most important, this class of algorithm typically works very well in practice.

In this paper, we provide large-sample optimality guarantees for a particular greedy single-edge search algorithm—called *greedy equivalence search* or *GES* for short—when that algorithm is used in conjunction with any *asymptotically consistent* scoring criterion. An asymptotically consistent scoring criterion is one that, in the limit of large number of samples, assigns the highest score to the *parameter-optimal* model–that is, the model with fewest parameters that can represent the generative distribution. It is well known that several Bayesian-network scoring criteria—including the Bayesian criterion and the minimum description length (MDL) criterion—are asymptotically consistent. We also provide optimality guarantees for an alternative greedy search algorithm that we call *unrestricted* GES.

Any greedy algorithm that is used in conjunction with an asymptotically consistent scoring criterion will (by definition) have a local maximum at a parameter-optimal model; the problem is that greedy search can get trapped in other local maxima. Furthermore, in order to apply the algorithm in practice, the connectivity of the search space must be sparse (i.e. the number of models considered at each step of the algorithm must be reasonably small). For example, because the number of models grows super-exponentially with the number of observed variables, the simple (greedy) approach of enumerating every model and selecting the best one is not realistic.

Meek (1997) shows that GES—under the assumption that "Meek's Conjecture" is true—will provably terminate at the parameter-optimal model whenever the generative distribution is *perfect* with respect to that model; that is, whenever the independence constraints among the observable variables in that distribution are precisely the independence constraints in the parameter-optimal model. This result is somewhat surprising because the connectivity of the search space for GES is sparse. Chickering (2002) proves that Meek's Conjecture is true—thus establishing the asymptotic optimality of GES—and provides an efficient implementation of the search operators used by GES. Furthermore, the experimental results of Chickering (2002) suggest that the large-sample guarantees of GES can hold with reasonably small sample sizes.

In this paper, we consider the large-sample behavior of GES when we eliminate the requirement that the generative distribution be perfect with respect to the parameter-optimal model. In particular, using a more realistic set of assumptions about the generative distribution, we show that the algorithm identifies a model that satisfies a weaker form of optimality that we call *inclusion optimality*. A model is inclusion optimal for a distribution if it can represent the distribution exactly and if no sub-model can also do so. Our results hold whenever the *composition* axiom of independence (see, e.g., Pearl, 1988) holds among the observable variables, the contrapositive of which states that whenever a variable $X$ is (conditionally or marginally) dependent on a set of variables $\mathbf{Y}$, then there is a singleton $Y \in \mathbf{Y}$ on which $X$ depends. A stronger but more intuitively appealing assumption that we make to guarantee that the composition holds is that the dependence relationships in the generative model can be characterized by paths between singleton nodes in some graphical model. Because the d-separation criterion identifies dependencies using paths of this type, it is easy to show that if the generative distribution is perfect with respect to some Bayesian network—where any subset of the nodes may be hidden—then we can guarantee inclusion optimality in the limit. Our results also apply when the generative distribution is perfect with respect to other types of graphical models including Markov random fields and chain graphs. In all of these situations, we allow for the presence of hidden variables in the generative model and selection bias in the observed data.

The paper is organized as follows. In Section 2, we describe our notation and previous relevant work. In

Section 3, we prove the main results of this paper. In Section 4, we describe a set of experiments we performed that demonstrate the practical importance of our results. Finally, in Section 5, we conclude with a summary and discussion of future relevant work.

## 2 Background

Throughout the paper, we use the following syntactical conventions. We denote a variable by an upper case letter (e.g. $A, B_i, Y, \Theta$) and a state or value of that variable by the same letter in lower case (e.g. $a, b_i, y, \theta$). We denote sets with bold-face capitalized letters (e.g. $\mathbf{A}, \mathbf{Pa_i}$) and corresponding sets of values by bold-face lower case letters (e.g. $\mathbf{a}, \mathbf{pa_i}$). Finally, we use calligraphic letters (e.g. $\mathcal{G}, \mathcal{B}$) to denote statistical models and graphs.

### 2.1 Directed Graphical models

In this paper, we concentrate on Bayesian networks for a set of variables $\mathbf{O} = \{X_1, \dots, X_n\}$, where each $X_i \in \mathbf{O}$ has a finite number of states. A *parametric Bayesian-network model* $\mathcal{B}$ for a set of variables $\mathbf{O} = \{X_1, \dots, X_n\}$ is a pair $(\mathcal{G}, \theta)$. $\mathcal{G} = (\mathbf{V}, \mathbf{E})$ is a directed acyclic graph—or *DAG* for short—consisting of (1) nodes $\mathbf{V}$ in one-to-one correspondence with the variables $\mathbf{O}$, and (2) directed edges $\mathbf{E}$ that connect the nodes. $\theta$ is a set of parameter values that specify all of the conditional probability distributions; we use $\theta_i \subset \theta$ to denote the subset of these parameter values that define the (full) conditional probability table of node $X_i$ given its parents in $\mathcal{G}$. A parametric Bayesian network represents a joint distribution over $\mathbf{O}$ that factors according to the structure $\mathcal{G}$ as follows:

$$p_{\mathcal{B}}(X_1 = x_1, \dots, X_n = x_n)$$
$$= \prod_{i=1}^{n} p(X_i = x_i | \mathbf{Pa_i}^{\mathcal{G}} = \mathbf{pa_i}^{\mathcal{G}}, \theta_i) \qquad (1)$$

where $\mathbf{Pa_i}^{\mathcal{G}}$ is the set of parents of node $x_i$ in $\mathcal{G}$. A *Bayesian-network model* (or DAG model) $\mathcal{G}$ is simply a directed acyclic graph and represents a family of distributions that satisfy the independence constraints that must hold in any distribution that can be represented by a parametric Bayesian network with that structure. We say that a Bayesian network $\mathcal{G}$ *includes* a distribution $p(\mathbf{O})$ if the distribution is defined by some parametric Bayesian network with structure $\mathcal{G}$.

The set of all independence constraints imposed by the structure $\mathcal{G}$ via Equation 1 can be characterized by the *Markov conditions*, which are the constraints that each variable is independent of its non-descendants given its parents. That is, any other independence constraint that holds can be derived from the Markov conditions (see, e.g., Pearl, 1988). Pearl (1988) provides a graphical condition called d-separation that can be used to identify any independence constraint that necessarily follows from the factorization. We use $A \perp\!\!\!\perp_{\mathcal{G}} B | \mathbf{S}$ to denote the assertion that DAG $\mathcal{G}$ imposes the constraint that $A$ is independent of $B$ given set $\mathbf{S}$.

## 2.2 Equivalence, Inclusion and Optimality

There are two common notions of equivalence for Bayesian networks. Bayesian networks $\mathcal{G}$ and $\mathcal{G}'$ are *distributionally equivalent* ($\mathcal{G} \approx_D \mathcal{G}'$) if for every parametric Bayesian network $\mathcal{B} = (\mathcal{G}, \theta)$, there exists a parametric Bayesian network $\mathcal{B}' = (\mathcal{G}', \theta')$ such that $\mathcal{B}$ and $\mathcal{B}'$ define the same probability distribution, and vice versa. Two DAGs $\mathcal{G}$ and $\mathcal{G}'$ are *independence equivalent* ($\mathcal{G} \approx_I \mathcal{G}'$) if the independence constraints in the two DAGs are identical. These two notions of equivalence are not generally the same, but they are for the parametric Bayesian-network models that we consider in this paper (i.e. the conditional distributions are specified with full tables) and thus we say that two DAGs $\mathcal{G}$ and $\mathcal{G}'$ are *equivalent*—denoted $\mathcal{G} \approx \mathcal{G}'$—to mean that they are both distributionally equivalent and independence equivalent.

Similarly, there are two corresponding types of inclusion relations for Bayesian networks. A Bayesian network $\mathcal{G}$ is *distributionally included* in a Bayesian network $\mathcal{H}$ ($\mathcal{G} \leq_D \mathcal{H}$) if every distribution included in $\mathcal{G}$ is also included in $\mathcal{H}$. A Bayesian network $\mathcal{G}$ is *independence included* in a Bayesian network $\mathcal{H}$ ($\mathcal{G} \leq_I \mathcal{H}$) if every independence relationship in $\mathcal{H}$ also holds in $\mathcal{G}$. The relationship $\mathcal{G} \leq_I \mathcal{H}$ is sometimes described in the literature by saying that $\mathcal{H}$ is *an independence map* of $\mathcal{G}$. If we assume that $\approx_I$ is equivalent to $\approx_D$ for a family of parametric Bayesian-network models then it is easy to show that $\leq_D$ is equivalent to $\leq_I$. Thus, because we are using complete tables, the two types of inclusion are equivalent and we use $\mathcal{G} \leq \mathcal{H}$ to denote that $\mathcal{G}$ is *included*—that is, both distributionally and independence—in $\mathcal{H}$. Note that we are using "included" to describe the relationship between a model and a particular distribution, as well as a relationship between two models. We say that $\mathcal{G}$ is *strictly included* in $\mathcal{H}$—denoted $\mathcal{G} < \mathcal{H}$—if $\mathcal{G}$ is included in $\mathcal{H}$ and $\mathcal{G}$ is not equivalent to $\mathcal{H}$.

In this paper, we are interested in two types of optimality. A Bayesian network $\mathcal{G}$ is *parameter optimal* for distribution $p$ if $\mathcal{G}$ includes $p$ and there is no Bayesian network that includes the distribution with fewer parameters. A Bayesian network $\mathcal{G}$ is *inclusion optimal* for distribution $p$ if $\mathcal{G}$ includes $p$ and there is no Bayesian network $\mathcal{G}'$ such that (1) $\mathcal{G}' \leq \mathcal{G}$ and (2) $\mathcal{G}'$ also includes $p$.

## 2.3 Learning Directed graphical models

Approaches to the Bayesian-network learning problem typically concentrate on identifying one or more Bayesian networks for a set of variables $\mathbf{O} = \{X_1, \ldots, X_n\}$ that best fit a set of observed data $\mathbf{D}$ for those variables according to some scoring criterion $S(\mathcal{G}, \mathbf{D})$; once the structure of a Bayesian network is identified, it is usually straightforward to estimate the parameter values for a corresponding (parametric) Bayesian network.

A scoring criterion $S(\mathcal{G}, \mathbf{D})$ is *score equivalent* if, for any pair of equivalent DAGs $\mathcal{G}$ and $\mathcal{H}$, it is necessarily the case that $S(\mathcal{G}, \mathbf{D}) = S(\mathcal{H}, \mathbf{D})$. A scoring criterion $S(\mathcal{G}, \mathbf{D})$ is *decomposable* if it can be written as a sum of measures, each of which is a function only of one node and its parents. In other words, a decomposable scoring criterion $S$ applied to a DAG $\mathcal{G}$ can always be expressed as:

$$S(\mathcal{G}, \mathbf{D}) = \sum_{i=1}^{n} s(X_i, \mathbf{Pa}_i^{\mathcal{G}}) \qquad (2)$$

Note that the data $\mathbf{D}$ is implicit in the right-hand side Equation 2. When we say that $s(X_i, \mathbf{Pa}_i^{\mathcal{G}})$ is only a function of $X_i$ and its parents, we intend this also to mean that the data on which this measure depends is restricted to those columns corresponding to $X_i$ and its parents.

Many commonly used scoring criteria are both score equivalent and decomposable. For a discussion of why score equivalence is an important (and sometimes necessary) property, see Heckerman, Geiger and Chickering (1995). One main advantage to using a decomposable scoring criterion is that if we want to compare the scores of two DAGs $\mathcal{G}$ and $\mathcal{G}'$, we need only compare those terms in Equation 2 for which the corresponding nodes have different parent sets in the two graphs. This proves to be particularly convenient for search algorithms that consider single edge changes.

To simplify the presentation in this paper, we concentrate on using the *Bayesian* scoring criterion, but emphasize that our results are more broadly applicable. For the Bayesian scoring criterion we define, for each model $\mathcal{G}$, a corresponding *hypothesis* $\mathcal{G}^h$, which for our purposes can simply denote the assertion that $\mathcal{G}$ is an inclusion-optimal representation of the generative distribution.[1] The scoring criterion is then defined to be the relative posterior (or relative log posterior) of $\mathcal{G}^h$ given the observed data. Without loss of generality, we express the Bayesian scoring criterion $S_B$ using the relative log posterior of $\mathcal{G}^h$:

$$S_B(\mathcal{G}, \mathbf{D}) = \log p(\mathcal{G}^h) + \log p(\mathbf{D}|\mathcal{G}^h) \qquad (3)$$

where $p(\mathcal{G}^h)$ is the prior probability of $\mathcal{G}^h$, and $p(\mathbf{D}|\mathcal{G}^h)$ is the *marginal likelihood*. The marginal likelihood is obtained by integrating the likelihood function (i.e. Equation 1) applied to each record in $\mathbf{D}$ over the unknown parameters of the model with respect to the parameter prior. Heckerman et al. (1995) describe parameter priors that guarantee score equivalence and score decomposability of the Bayesian criterion.

---

[1]In practice, the definition of DAG hypothesis is important only to the extent in which it determines how the second term is evaluated in Equation 3. For most definitions found in the literature, the resulting values for this term are identical.

## 2.4 Asymptotically Consistent Scores

It is well known that the Bayesian scoring criterion is *asymptotically consistent*. Simply stated, an asymptotically consistent scoring criterion is one that—in the limit as the number of observed cases grows large—prefers the model containing the fewest number of parameters that can represent the generative distribution exactly. Geiger, Heckerman, King and Meek (2001) show that parametric Bayesian-network models that contain complete tables are curved exponential models; Haughton (1988) derives the following approximation for the Bayesian criterion for this model class:

$$S_B(\mathcal{G}, \mathbf{D}) = \log p(\mathbf{D}|\hat{\theta}, \mathcal{G}^h) + \frac{d}{2}\log m + O(1) \qquad (4)$$

where $\hat{\theta}$ denotes the maximum likelihood values for the network parameters, $d$ is the dimension of the model and $m$ is the number of records in $\mathbf{D}$. The first two terms in this approximation are known as the *Bayesian information criterion* (or BIC). The presence of the $O(1)$ error means that, even as $m$ approaches infinity, the approximation can differ from the true relative log posterior by a constant. As shown by Haughton (1988), however, BIC is consistent. Furthermore, it can be shown that the leading term in BIC grows as $O(m)$, and therefore we conclude that because the error term becomes increasingly less significant as $m$ grows large, Equation 3 is consistent as well. Because the prior term $p(\mathcal{G}^h)$ does not depend on the data, it does not grow with $m$ and therefore is absorbed into the error term of Equation 4. Thus the asymptotic behavior of the Bayesian scoring criterion depends only on the marginal likelihood term.

Consistency of the Bayesian scoring criterion leads, from the fact that BIC is decomposable, to a more useful property of the criterion that we call *local consistency*. Intuitively, if a scoring criterion is locally consistent, then the score of a DAG model $\mathcal{G}$ (1) *increases* as the result of adding any edge that eliminates an independence constraint that does not hold in the generative distribution, and (2) *decreases* as a result of adding any edge that does not eliminate such a constraint. More formally, we have the following definition.

## Definition (Local Consistency)

*Let $\mathbf{D}$ be a set of data consisting of $m$ records that are iid samples from some distribution $p(\cdot)$. Let $\mathcal{G}$ be any DAG, and let $\mathcal{G}'$ be the DAG that results from adding the edge $X_j \to X_i$. A scoring criterion $S(\mathcal{G}, \mathbf{D})$ is locally consistent if, in the limit of large $m$, the following two properties hold:*

1. *If $X_j$ is not independent of $X_i$ given $\mathbf{Pa}_i^{\mathcal{G}}$ in $p$, then $S(\mathcal{G}', \mathbf{D}) > S(\mathcal{G}, \mathbf{D})$*
2. *If $X_j$ is independent of $X_i$ given $\mathbf{Pa}_i^{\mathcal{G}}$ in $p$ then $S(\mathcal{G}', \mathbf{D}) < S(\mathcal{G}, \mathbf{D})$*

Chickering (2002) shows that the Bayesian scoring criterion is locally consistent, a result we present formally below.

**Lemma 1 Chickering (2002)** *The Bayesian scoring criterion is locally consistent.*

The significance of Lemma 1 is that as long as there are edges that can be added to a DAG that eliminate independence constraints not contained in the generative distribution, the Bayesian scoring criterion will favor such an addition. Furthermore, if the generative distribution is included in a DAG, then Lemma 1 guarantees that any deletion of an "unnecessary" edge will be favored by the criterion.

## 2.5 Greedy Equivalence Search

In this section, we describe the greedy single-edge search algorithm that we use for learning Bayesian networks. Rather than searching over the space of DAGs, we use equivalence classes of DAGs defined by the (reflexive, symmetric, and transitive) equivalence relation $\approx$ defined in Section 2.2. We use $\mathcal{E}$ to denote an equivalence class of DAG models. Note that we use the *non-bold* character $\mathcal{E}$; although arguably misleading in light of our convention to use bold-face for sets of variables, we use the non-bold character to emphasize the interpretation of $\mathcal{E}$ as a model for a set of independence constraints as opposed to a set of DAGs. To denote a particular equivalence class to which a DAG model $\mathcal{G}$ belongs, we sometimes write $\mathcal{E}(\mathcal{G})$. Note that $\mathcal{G} \approx \mathcal{G}'$ implies $\mathcal{G}' \in \mathcal{E}(\mathcal{G})$ and $\mathcal{G} \in \mathcal{E}(\mathcal{G}')$. We extend the definition of inclusion to pertain to equivalence classes of DAGs in the obvious way.

The connectivity of the search space is defined using the inclusion relation ($\leq$) between DAGs. In particu-lar, two equivalence classes $\mathcal{E}_1(\mathcal{G})$ and $\mathcal{E}_2(\mathcal{G}')$ are adjacent if and only $G \leq G'$ or $G' \leq G$ and the number of edges in the graphs $G$ and $G'$ differ by one. We say we are moving in the search space in a *forward direction* if we move from a state $\mathcal{E}_1(\mathcal{G})$ to an adjacent state $\mathcal{E}_2(\mathcal{G}')$ in which $G \leq G'$; otherwise we are moving in a *backward direction*.

The *greedy equivalence search* algorithm (or GES for short) is a two-phase greedy algorithm that can be described as follows. The algorithm starts with the equivalence class corresponding to no dependencies among the variables (i.e. the class containing the DAG model with no edges). Then, for the first phase, a greedy search is performed only in the forward direction until a local maximum is reached. For the second phase, a second greedy search is performed, starting from the local maximum from the first phase, but this time only in the backward direction. GES terminates with the local maximum reached by the second phase. We find it convenient to name the (restricted) greedy searches in the first and second phase of GES *forward equivalence search* (FES for short) and *backward equivalence search* (BES for short), respectively. GES can thus be described as running FES starting from the all-independence model, and then running BES starting from the resulting local maximum.

GES is a restricted version of a more general greedy search algorithm that considers moves in both the forward and backward directions at each step. We call this unrestricted version of the search *unrestricted GES* or *UGES* for short.

For states of the search in which the dependency structure of the equivalence class is very dense, the number of adjacent states that need be considered by GES and UGES can be exponential in the number of variables. For simple models, however, the number of adjacent states is small. Fortunately, we have found that in practice the algorithms—when applied to real data sets—only encounter simple models. In fact, Chickering (2002) demonstrates that GES is as fast as a greedy DAG-based search that considers $O(n^2)$ adjacent states at each step.

Chickering (2002) describes a representation for equivalence classes and a corresponding set of operators that implement the forward and backward searches of GES. All of the operators can be identified efficiently and can be scored—when using a decomposable scoring criterion—by evaluating only a small subset of the terms in Equation 2.

We end this section by presenting a transformational characterization of the inclusion relation for Bayesian networks. The characterization was initially conjectured to be valid by Meek (1997), and was later proven to be so by Chickering (2002).

**Theorem 2 (Chickering, 2002)** *Let $\mathcal{G}$ and $\mathcal{H}$ be any pair of DAGs such that $\mathcal{G} \leq \mathcal{H}$. Let $r$ be the number of edges in $\mathcal{H}$ that have opposite orientation in $\mathcal{G}$, and let $a$ be the number of edges in $\mathcal{H}$ that do not exist in either orientation in $\mathcal{G}$. There exists a sequence of at most $r + 2a$ distinct edge reversals and additions in $\mathcal{G}$ with the following properties:*

1. *Each edge reversed is a covered[2] edge*
2. *After each reversal and addition $\mathcal{G}$ is a DAG and $\mathcal{G} \leq \mathcal{H}$*
3. *After all reversals and additions $\mathcal{G} = \mathcal{H}$*

This theorem plays an essential role to understanding how the GES algorithm—and more specifically the BES algorithm—leads to an inclusion-optimal model. The key feature in the characterization is that it is based on single edge transformations.

## 3 Results

In this section, we prove the main results of this paper. Throughout the section, we use $p$ to denote the distribution over the observable variables from which the observed data $\mathbf{D}$ was generated, and we use $m$ to denote the number of records in $\mathbf{D}$.

First we show that the second phase of the algorithm (i.e. the BES algorithm) is guaranteed (in the limit of large $m$) to identify an inclusion-optimal model if it starts with an equivalence class that includes $p$.

**Theorem 3** *If $\mathcal{E}^*$ includes $p$ then, in the limit of large $m$, the result of running BES, starting from $\mathcal{E}^*$ and using any locally consistent scoring criterion, results in an inclusion-optimal model.*

**Proof:** After each step in the backward equivalence search, we are guaranteed that the current state $\mathcal{E}$ will include $p$ by the following argument. Suppose this is not the case, and consider the first move made by BES to a state that does not include $p$. Because this move corresponds to an edge deletion in some DAG, it follows immediately from the fact that the scoring function is locally consistent that any such deletion would *decrease* the score, thus contradicting the fact that BES is greedy.

To complete the proof, assume that BES terminates with some equivalence class $\mathcal{E}$ that is not inclusion optimal, and let $\mathcal{E}' < \mathcal{E}$ be any inclusion optimal equivalence class that is strictly included in $\mathcal{E}$. Let $\mathcal{H}$ be any DAG in $\mathcal{E}$, and let $\mathcal{G}$ be any DAG in $\mathcal{E}'$. Because $\mathcal{G} < \mathcal{H}$ we conclude from Theorem 2 that there exists a sequence of covered edge reversals and edge additions that transforms $\mathcal{G}$ into $\mathcal{H}$. There must be at least one edge addition in the sequence because by assumption $\mathcal{G} \not\approx \mathcal{H}$ and because (see Chickering, 1995) reversing a covered edge in a DAG results DAG in the same equivalence class. Consider the DAG $\mathcal{G}'$ that precedes the *last* edge addition in the sequence. Clearly $\mathcal{E}(\mathcal{G}')$ is one step backwards from $\mathcal{E}$ and because $\mathcal{G}'$ has fewer parameters than $\mathcal{H}$, we conclude from the local consistency of the scoring criterion that $\mathcal{E}$ cannot be a local minimum, yielding a contradiction. $\square$

---

[2]An edge $X_i \rightarrow X_j$ is covered in DAG $\mathcal{G}$ if $\mathbf{Pa}_j^{\mathcal{G}} = \mathbf{Pa}_i^{\mathcal{G}} \cup X_i$.

Theorem 3 is important from a theoretical point of view because we can always start BES with the *complete* equivalence class that asserts no independence constraints; this model is guaranteed to include $p$. One problem with starting from the complete model is that for any realistic domain, the number of parameters in the model will be prohibitively large. Put another way, in order for the asymptotic properties of the algorithm to apply to a real (finite $m$) problem, we would need an unrealistic number of records in the data. Another problem with starting from the complete model is that BES must begin by evaluating an exponential number of (adjacent) states, thus making the algorithm intractable in this situation. As we shall see in Theorem 4, the previous theorem becomes important in a practical sense if the *composition* property holds in $p$ among the observable variables: given that a variable $X$ is not independent of the set $\mathbf{Y}$ given set $\mathbf{Z}$, then there exists a singleton element $Y \in \mathbf{Y}$ such that $X$ is not independent of $Y$ given set $\mathbf{Z}$.

**Theorem 4** *If $p$ satisfies the composition property then, in the limit of large $m$, GES using any locally consistent scoring criterion finds an inclusion optimal model.*

**Proof:** Given Theorem 3 we need only to show that the forward search (FES) in the first phase of GES identifies an equivalence class that includes $p$. Suppose this is not the case, and consider any DAG $\mathcal{G}$ contained in the (local maximum) equivalence class reached at the end of the first phase of GES. Because $\mathcal{G}$ does not include $p$, there must be some independence constraint from $\mathcal{G}$ that does not hold in $p$. Because the independence constraints of $\mathcal{G}$ are characterized by the Markov conditions, it follows that in $p$, there must exist some node $X_i$ in $\mathcal{G}$ for which $X_i$ is not independent of its non-descendants $\mathbf{Y}$ given its parents $\mathbf{Pa}_i$. Because the composition axiom holds for $p$, there must exist at least one singleton non-descendant $Y \in \mathbf{Y}$ for which this dependence holds. By Lemma 1, this implies that the DAG $\mathcal{G}'$ that results from adding the edge $Y \rightarrow X_i$ to $\mathcal{G}$ (which cannot be cyclic by definition of $\mathbf{Y}$) has a higher score than $\mathcal{G}$. The equivalence class $\mathcal{E}(\mathcal{G}')$ is one step forward from $\mathcal{E}$ which contradicts the fact that $\mathcal{E}$ is a local maximum. $\square$

The proof of Theorem 4 does not require that GES start with the empty (i.e. no-dependence) model; as a result, we obtain the following corollary.

**Corollary 5** *If $p$ satisfies the composition property then, in the limit of large $m$, UGES using any locally consistent scoring criterion finds an inclusion optimal model.*

**Proof:** Suppose the corollary is not correct. Then there exists a local maximum $\mathcal{G}$ in the UGES search space that is not inclusion optimal. From the proof of Theorem 4, we can run GES starting from the model $\mathcal{G}$ to reach an inclusion-optimal model. Because the operators available to GES are a strict subset of the operators in UGES, it follows that $\mathcal{G}$ cannot be a local maximum in the UGES search space, yielding a contradiction. $\square$

Theorem 4 and Corollary 5 are very general results in the sense that we assume nothing about $p$ except that the composition property holds over the observable variables; the composition property need not hold among any of the variables involved that are not observed. The "composition assumption" in isolation, however, may not be intuitively appealing to many. In what situations is this assumption violated? Can we expect the composition assumption to be reasonable in many domains?

To help gain a better understanding of the types of situations for which the composition property holds, we introduce the notion of a *graphical path condition*. A graphical path condition $PC_{\mathcal{G}}(X, Y, \mathbf{Z})$ is a function of a graphical model $\mathcal{G}$ that maps two singleton nodes and a set of nodes to either zero or one. Intuitively, the function checks whether or not there is a "path" from $X$ to $Y$ given "context" $\mathbf{Z}$. The d-separation criterion, for example, is a graphical path condition for DAG models: in this case $PC_{\mathcal{G}}(X, Y, \mathbf{Z})$ has the value one if and only if there is an active path from $X$ to $Y$ in $\mathcal{G}$ given set $\mathbf{Z}$. As another example, the presence of an undirected path between $X$ and $Y$ that does not pass through a node in $\mathbf{Z}$ is a graphical path condition for a Markov random field (undirected graphical model). As we discuss below, when there exists a graphical path condition that characterizes the dependencies among the variables—as is the case with both of the previous examples—then we are guaranteed that the composition property will hold. To simplify the discussion, we provide the following definition.

**Definition (Path Property)**

*A graphical model $\mathcal{M}$ has the* path property *if there exists a path condition $PC_{\mathcal{M}}(\cdot, \cdot, \cdot)$ that characterizes the dependencies implied by $\mathcal{M}$ as follows:*

$$\mathbf{X} \not\perp\!\!\!\perp_{\mathcal{M}} \mathbf{Y} | \mathbf{Z} \Leftrightarrow \exists X \in \mathbf{X}, Y \in \mathbf{Y} \, s.t. PC_{\mathcal{M}}(X, Y, \mathbf{Z}) = 1$$

We now show that if the generative distribution is perfect with respect to a model that has the path property, then the composition axiom holds even in the presence of hidden variables and *selection variables*. Selection variables are hidden variables that are in a particular state for each record in the observed data. In a mail survey, for example, a selection variable might correspond to "the person filled out the survey and mailed it back"; the presence of such variables can lead to biased results because those who respond to the survey may not be representative of the population as a whole.

Let a $\mathcal{M}$ be a graphical model for variables $\mathbf{V} = \{\mathbf{O} \cup \mathbf{H} \cup \mathbf{S}\}$ where $\mathbf{O}$ is a set of observed variables, $\mathbf{H}$ a set of hidden variables and $\mathbf{S}$ a set of selection variables.

**Proposition 1** *If $q$ is a distribution that is perfect with respect to a model $\mathcal{M}$ that has the path property, then the composition property holds for $p(\mathbf{O}) = \sum_{\mathbf{H}} q(\mathbf{O}, \mathbf{H}, \mathbf{S} = \mathbf{s})$.*

**Proof:** Follows immediately from the definition of the path property and from the fact that because $q$ is perfect with respect to $\mathcal{M}$, $q$ and $\mathcal{M}$ have precisely the same dependence relations. □

This proposition tells us that if our data is generated from a distribution that is perfect with respect to some graphical model with the path property then the distribution is guaranteed to satisfy the composition property even if there are hidden variables and the data is generated with selection bias. This naturally leads to the next corollary.

**Corollary 6** *If $p$ is perfect with respect to either a DAG, a Markov random field, or a chain graph, then in the limit of large $m$, GES (or UGES) using any locally consistent scoring criterion finds an inclusion optimal model.*

**Proof:** This follows immediately because all of these types of graphical models have the path property. □

# 4 Experiments

In this section, we present experimental results demonstrating that we can attain the large-sample benefits of the GES algorithm—that is, we can identify the inclusion-optimal model—when the generative distribution is not DAG-perfect and when given a finite sample size. Our approach is to sample data from known *gold-standard* models for which we can analytically determine the inclusion-optimal models defined over the observable variables. By generating synthetic data from the gold-standard models, we can evaluate how well GES performs by checking whether or not it identifies the corresponding inclusion-optimal model.

We concentrate on two specific gold-standard structures for all of our experiments; we sample the corresponding generative parameters—using a random-sampling technique described below—to produce the generative distribution. The first gold-standard structure is the *w-structure model* shown in Figure 1a in which all variables are binary except for the three-valued $X_2$; the variable $H$ is hidden, and as a result there is no perfect map in a DAG model defined over the observables. The second generative structure we consider is the *selection four-cycle model* shown in Figure 2a in which all variables are binary except for the four-valued $X_1$. The variable $S$ is a selection variable with a corresponding selection value of one; in other words, given a random sample of cases from this model, we only allow ones for which $S = 1$ to be included in the observed data. The resulting distribution over the observable variables is included in an undirected four cycle, which has no perfect map in a DAG model.

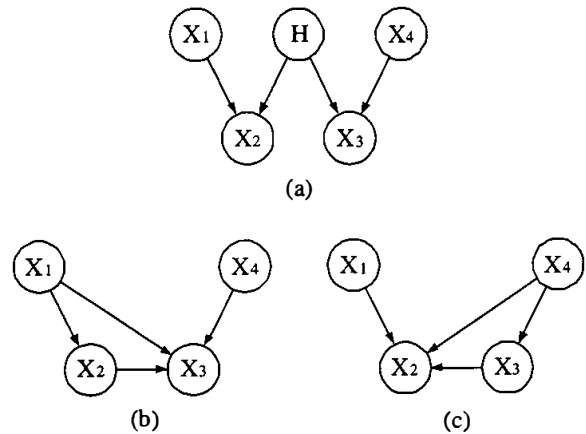

(a)

(b)                          (c)

Figure 1: (a) the w-structure model, (b) the parameter-optimal model and (c) an inclusion-optimal model that is not parameter optimal.

In Figure 1b and Figure 1c we show (the unique) representative DAG models from the two inclusion-optimal equivalence classes corresponding to the w-structure model. The model in Figure 1b, which contains 18 parameters, is parameter optimal, whereas the model in Figure 1c, which contains 20 parameters, is not pa-rameter optimal. Similarly, in Figure 2b and Figure 2c we show representative DAG models from the two inclusion-optimal equivalence classes corresponding to the selection four-cycle model. The model in Figure 2b, which contains 19 parameters, is parameter optimal, whereas the model in Figure 2b, which contains 23 parameters, is not parameter optimal.

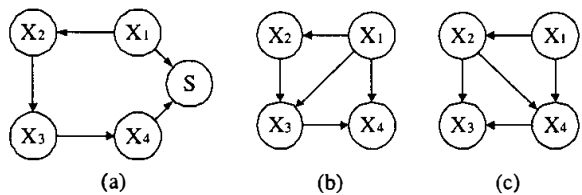

Figure 2: (a) the selection four-cycle model, (b) the parameter-optimal model and (c) an inclusion-optimal model that is not parameter optimal.

In order to produce a random generative distribution with strong dependencies between the variables, we sampled each of the conditional parameter distributions from the gold standard as follows. For a variable $X_i$ with $k$ states, we constructed a "basis" mean value $\vec{\mu}$ for $p(X_i|\mathbf{Pa}_i^{\mathcal{G}})$ by normalizing the vector $(\frac{1}{1}, \frac{1}{2}, \ldots, \frac{1}{k})$. For the $j$th "instantiation" $\mathbf{pa}_i^{\mathcal{G}}$ of $\mathbf{Pa}_i^{\mathcal{G}}$ we produced the mean value $\vec{\mu}_j$ by shifting $\vec{\mu}$ to the right $j$ places when $j$ modulo $k$ was not one. For example, if $\vec{\mu} = \alpha \cdot (1, \frac{1}{2}, \frac{1}{3})$ (where $\alpha$ is the normalization constant), then $\vec{\mu_1} = \alpha \cdot (\frac{1}{3}, 1, \frac{1}{2})$, $\vec{\mu_2} = \alpha \cdot (\frac{1}{2}, \frac{1}{3}, 1)$, and so on. We then sampled $p(X_i|\mathbf{Pa}_i^{\mathcal{G}} = \mathbf{pa}_i^{\mathcal{G}})$ from a Dirichlet distribution with mean $\vec{\mu_j}$ and equivalent sample size of 10. The choice of this prior distribution for the conditional parameters ensures a reasonable level of dependence between d-connected variables in the generative structure.

Our experiments proceeded as follows. We considered 17 sample sizes, starting with $m = 10$ and then doubling to obtain the next sample size until $m = 655360$. For each sample size, we produced 100 random generative distributions for both of the generative structures. From each such generative distribution, we sampled a single data set of the appropriate size that contained only those values for the observable variables $\{X_1, X_2, X_3, X_4\}$. For the selection four-cycle model, any sample in which $S$ was not in the selection state was discarded; samples were taken from this model until the number of non-discarded records was equal to $m$. We then ran the GES algorithm using the BDeu scoring criterion (described by Heckerman et al., 1995) with a uniform structure prior and an equivalent sample size of ten. In particular, the version of the crite-

rion that we used can be expressed as:

$$S(\mathcal{G}, \mathbf{D}) = \log \prod_{i=1}^{n} \prod_{j=1}^{q_i} \frac{\Gamma(\frac{10}{q_i})}{\Gamma(\frac{10}{q_i} + N_{ij})} \cdot \prod_{k=1}^{r_i} \frac{\Gamma(\frac{10}{r_i \cdot q_i} + N_{ijk})}{\Gamma(\frac{10}{r_i \cdot q_i})}$$

where $q_i$ is the number of configurations of $\mathbf{Pa}_{X_i}^{\mathcal{G}}$, $r_i$ is the number of configurations (states) of $X_i$, $N_{ijk}$ is the number of records in the data for which $X_i = k$ and $\mathbf{Pa}_{X_i}^{\mathcal{G}}$ is in the $j$th configuration, and $N_{ij} = \sum_k N_{ijk}$. $\Gamma(\cdot)$ is the *Gamma* function, which satisfies $\Gamma(y+1) = y\Gamma(y)$ and $\Gamma(1) = 1$. Finally, after each run of GES, we compared the resulting local maximum to the corresponding inclusion-optimal model(s).

Figure 3a and Figure 3b show the results of our experiments corresponding to the w-structure and selection four-cycle model, respectively. In these figures we record, for each sample size, the percentage of models identified by GES that are inclusion optimal. As expected, as the sample size increases, the algorithm is more likely to identify the optimal model.

As discussed above, corresponding to each of the domains are two different equivalence-classes of models that are inclusion optimal. Only one of these classes is parameter optimal, however, and the heights of the curves in Figure 3 are the sum of (1) the percent of parameter-optimal models and (2) the percent of inclusion-optimal models that are not parameter optimal. When we broke these sums into their component parts, we found that for all sample sizes—and for both domains—GES identifies the parameter-optimal model in roughly a constant portion of those times when it identifies an inclusion-optimal model. In particular, the algorithm identified the parameter-optimal model roughly three fourths of the time for the w-structure model, and roughly half the time for the selection four-cycle model, regardless of the sample size. These results suggest that even in the large-sample limit, GES may not be able to reliably identify the parameter-optimal model.

## 5 Conclusion and Final Remarks

In this paper, we proved that in the limit of large sample sizes, the GES algorithm identifies an inclusion-optimal equivalence class of DAG models. The result is an important extension to the results of Meek (1997) and Chickering (2002) because—although it guarantees a weaker form of optimality—it relaxes the assumption that the generative distribution is DAG-perfect among the observable variables. Our results instead rely on the composition property of independence holding among the observable variables. This weaker assumption necessarily holds whenever the generative distribution is perfect with respect to a model that has the path property (a more reasonable assumption), regardless of whether that model contains hidden variables or whether the observed data is biased from hidden selection variables.

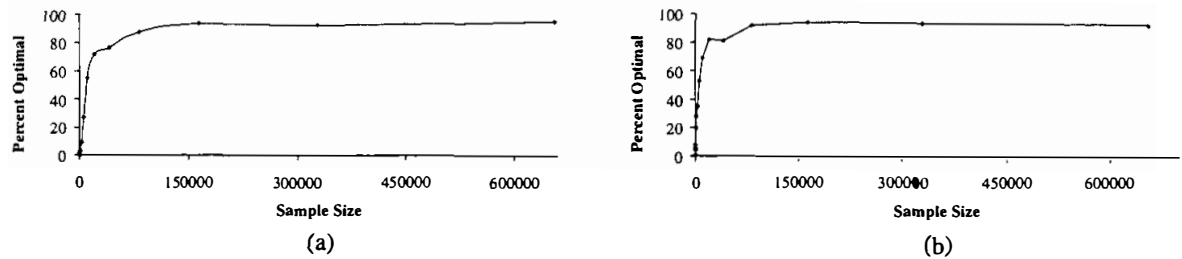

Figure 3: Percentage of models identified by GES that are inclusion optimal as a function of the sample size for (a) the w-structure gold standard and (b) the selection four-cycle gold standard.

When the generative distribution is DAG-perfect among the observable variables, there is a unique inclusion-optimal model that is identical to the (unique) parameter-optimal model. As we saw in Section 3, however, when the generative distribution is not DAG-perfect among the observable variables, there can be multiple inclusion-optimal models, some of which are not parameter optimal. Furthermore, there may be more than one parameter-optimal model. Our experiments suggest that GES may not be able to identify a parameter-optimal model, even in the limit of large sample size. An interesting area for further investigation is to identify a set of general conditions under which greedy algorithms (e.g. GES and UGES) will identify the parameter-optimal model.

Our experiments showed that GES can identify inclusion-optimal models when given large datasets. More work should be done to compare GES to other algorithms (such as UGES) while varying the sample size; alternative search algorithms might perform better with small datasets. Finally, it would be useful to investigate further the number of samples required to obtain reliably an inclusion-optimal model using algorithms such as GES; Chickering and Meek (2002) provide additional empirical evidence suggesting that this number can be large.